%% file: main.tex
\documentclass[11pt]{article}

\usepackage[preprint]{acl}

\usepackage{times}
\usepackage{latexsym}
\usepackage{svg}
\usepackage{enumitem}
\usepackage{multirow}

\usepackage{lipsum}

\usepackage[T1]{fontenc}

\usepackage[utf8]{inputenc}
\usepackage{microtype}
\usepackage{amsmath}
\usepackage{amssymb}
\usepackage{inconsolata}
\usepackage{graphicx}
\usepackage{cuted}
\usepackage{caption}
\usepackage{svg}
\usepackage{booktabs}
\usepackage{makecell}
\usepackage{subcaption} % in preamble

\DeclareMathOperator*{\argmin}{arg\,min}

\newcommand{\module}[1]{\textsc{#1}}
\newcommand{\feature}[1]{\texttt{#1}}
\newcommand{\model}[1]{\textit{#1}}

\definecolor{darkgreen}{rgb}{0,0.39,0}
\newif\ifshowedits
\showeditstrue

\title{Improving Medical Communication using Rubric-Guided Counterfactual Recommendations}

\author{
 \textbf{Adrian Cosma\textsuperscript{\S}},
 \textbf{Nicoleta-Nina Basoc\textsuperscript{$\dagger$}},
 \textbf{Andrei Niculae\textsuperscript{$\dagger$}},
\\
 \textbf{Cosmin Dumitrache\textsuperscript{$\dagger$}},
 \textbf{Emilian Radoi\textsuperscript{$\dagger$}}
\\
\\
 \textsuperscript{\S}IDSIA, Dalle Molle Institute for Artificial Intelligence,\\
 \textsuperscript{$\dagger$}National University of Science and Technology POLITEHNICA Bucharest,
\\
 \small{
   \textbf{Correspondence:} \href{mailto:emilian.radoi@upb.ro}{emilian.radoi@upb.ro}
 }
}

\begin{document}
\maketitle

\begin{abstract}
Text-based telemedicine increasingly relies on lightweight patient feedback, however, such feedback primarily reflects perceived communication quality rather than medical accuracy. We introduce an LM-guided counterfactual recommendation pipeline that discovers and refines interpretable communication features  such as tone, personalization, actionability and completeness in addressing patient concerns, without interfering with the medical content. These features are used together with patient-doctor interaction metadata to estimate positive feedback. At inference time, the system searches over low-cost ordinal feature changes and recommends minimal communication changes predicted to increase the probability of positive feedback, while independent auditor models test whether these gains generalize beyond the selection model. Across interactions, recommendations yield a mean +6.41\% gain in predicted positive feedback probability under independent auditors, and are non-negative for 93.31\% of recommendations. These results suggest that small, interpretable communication changes can capture most predicted gains while preserving the doctor's control over medical reasoning and final wording.
\end{abstract}

\section{Introduction}
\begin{figure*}[hbt!]
    \centering
    \includesvg[width=0.85\linewidth]{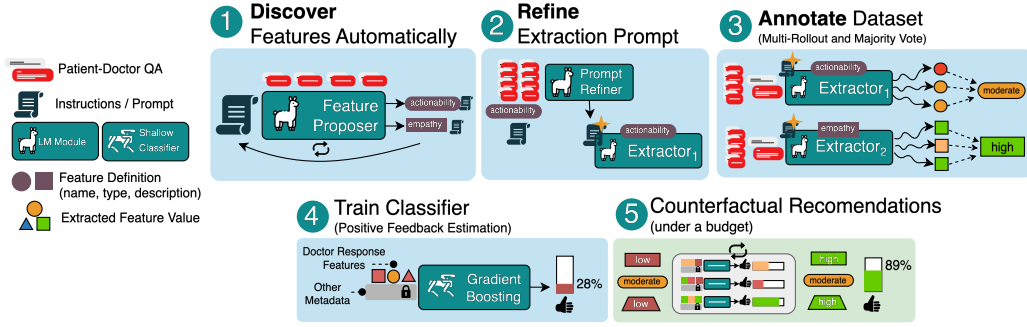}
    \caption{\textbf{Overview of the proposed pipeline.} Starting from patient--doctor QA pairs, an LM first optimizes a set of interpretable response-level features, which are then refined into feature-specific extraction prompts and used to annotate the dataset. The extracted semantic features, together with structured metadata, are used to train a feedback prediction model. At inference time, we search over counterfactual changes to the semantic features and recommend the smallest feature modifications that maximize the probability of positive feedback.}
    \label{fig:diagram}
\end{figure*}

\input{sections/1.intro}

\section{Related Work}
\input{sections/2.related}

\section{Method}
\input{sections/3.method}

\section{Experiments and Results}

\input{sections/4.exps-results}

\section{Conclusions}
\input{sections/5.conclusions}

\section*{Limitations}
\input{sections/6.limitations}

\bibliography{refs}

\appendix
\input{sections/7.appendix}

\end{document}

%% file: sections/1.intro.tex
Telemedicine platforms increasingly mediate the first point of contact between patients and doctors. In 2024, there were reportedly over 116 million users of online consultations worldwide \cite{statista}. Some of these interactions occur through written text, where patients submit questions, doctors respond asynchronously, and the exchange typically concludes with a lightweight feedback signal such as a positive or negative rating. This feedback is important, as it shapes doctors' reputations, serves as a proxy for patient satisfaction, and incentivizes doctors to communicate more effectively with patients.

However, patient feedback does not reflect the medical accuracy of responses, as patients are not well positioned to evaluate clinical correctness. Their judgments are mostly driven by communication qualities such as clarity, empathy, personalization and actionability. 
Prior work has shown that doctor communication significantly influences patient experience, treatment adherence and perceptions of care \citep{martin2005challenge,geng2024association}. This raises an important question for telemedicine platforms: \textit{how can doctors be supported in improving the patient-facing quality of their responses without automating medical advice or rewriting their response?} 

We address this problem as one of \emph{counterfactual recommendation}. Given a patient question, a doctor response, and question-response metadata, our goal is to identify the smallest set of communication features of the response that, if improved, would most increase the probability of positive patient feedback. Unlike previous works \cite{niculae-etal-2025-dr,li2023chatdoctor,zhao2025smart,zhao2025medrag}, we do not use a \emph{Language Model} (LM) to generate a new medical response. We recommend interpretable editing targets, such as increasing personalization, providing clearer explanations, or making recommendations more actionable. The doctor retains full control over the final response and may accept, modify or ignore the recommendations.

This setting imposes three \textit{requirements} on any system for improving doctor responses: 

\begin{description}
    \item[$\mathfrak{R}_1$ \textbf{Explainability}:] Recommendations should be expressed in terms of interpretable communication features.
    \item[$\mathfrak{R}_2$ {\textbf{Feedback Improvement}}:] Recommended changes should increase the predicted probability of positive patient feedback.
    \item[$\mathfrak{R}_3$ {\textbf{Minimal Editing Effort}}:] Recommendations should achieve feedback improvement with minimal editing effort.
    % while requiring the fewest possible modifications.
\end{description}

To satisfy these requirements, we propose an LM-guided counterfactual recommendation pipeline for text-based telemedicine responses, that leverages \emph{Automatic Prompt Optimization} (APO) techniques. 

This paper makes the following contributions:
\begin{enumerate}
    \item We adapt and validate automated feature discovery \cite{cosma2026automaticpromptoptimizationdatasetlevel} for text-based telemedicine feedback modeling, using LMs to discover, refine and extract response-level semantic features predictive of patient satisfaction ($\mathfrak{R}_1$, $\mathfrak{R}_2$). We evaluate the predictive value of these features by training an interpretable feedback estimator which achieves 71.5\% ROC-AUC.
    \item We propose a budget-constrained counterfactual search procedure over ordinal semantic features, generating low-cost recommendations by balancing predicted feedback gains against editing effort ($\mathfrak{R}_2$, $\mathfrak{R}_3$). We perform a quantitative study on the proposed recommendations, and find that they raise the predicted probability of positive feedback for 93.31\% of responses, by +6.41\% on average.
\end{enumerate}

%% file: sections/2.related.tex
\noindent \textbf{LM-Driven Feature Discovery.}
While LMs have been used in the past for reliable feature \emph{extraction} under a fixed schema \cite{he2024annollm,gilardi2023chatgpt,tornberg2023chatgpt4outperformsexpertscrowd}, recent works have used LMs to \emph{discover} what features to extract in the first place \citep{cosma2026automaticpromptoptimizationdatasetlevel,balek2025llm,zhou-etal-2024-llm-feature,zhang2025multiagent}. Beyond optimizing instructions for a single prediction, as in APO \citep{zhou2022large,pryzant-etal-2023-automatic,yuksekgonul2025optimizing},  \citet{cosma2026automaticpromptoptimizationdatasetlevel} frame feature discovery as a dataset-level prompt optimization problem, where the optimized prompt induces a shared feature schema scored by downstream classifier performance. We adapt and improve upon this framework by incorporating the unsupervised prompt refinement step, and apply it to telemedicine feedback. We treat the discovered features as the action space for counterfactual recommendations \cite{jiang2024robust}. 

\noindent \textbf{Communication Quality in Medicine.}
Clinician communication shapes patient adherence, health outcomes, and how patients judge their doctors \citep{martin2005challenge,street2009does,stergiopoulos2023makes}. On telemedicine platforms, the linguistic characteristics of doctor responses are associated with patient satisfaction \citep{geng2024association}. Although LMs are increasingly used in healthcare \citep{anthropic-economic-index}, patient-facing deployment remains difficult: users assisted by LMs perform no better than controls \citep{bean2025reliability}, and advice believed to involve AI is trusted less \citep{reis2024influence,hohenstein2023artificial}. This motivates doctor-facing assistance that preserves the doctor as the author.

%% file: sections/3.method.tex
% We propose a doctor-in-the-loop system for recommending minimal, interpretable changes to telemedicine doctor responses. The system does not generate or rewrite medical advice; it identifies communicative aspects of a response that are associated with positive patient feedback and recommends which of them to improve, leaving the doctor responsible for the final response.

Our pipeline has five stages, as shown in Fig. \ref{fig:diagram}. We first apply dataset-level automatic feature discovery \citep{cosma2026automaticpromptoptimizationdatasetlevel} to optimize a set of response-level semantic features that are likely to be predictive of patient feedback. Each feature describes a mutable communication property of the doctor's response. We then refine feature-specific extraction prompts using a grounded prompt refinement procedure inspired by APO methods such as MIPRO \citep{opsahl2024optimizing}. These extractors annotate each patient--doctor interaction with ordinal semantic feature values. 

The extracted semantic features, together with structured question-response metadata, are used to train a feedback estimator that predicts the probability of positive patient feedback. At inference time, we enumerate low-cost positive changes to the semantic representation of a response and select the counterfactual that best trades off predicted feedback improvement against editing effort. The output is a small set of feature-level recommendations describing which communication aspects should be improved.

Consider a patient who asks \textit{``I have had headaches almost every day for the past two weeks. Should I be worried?''}, and a doctor who responds \textit{``It is probably a tension headache. I recommend a consultation with a neurologist''}. The response is medically reasonable, but it is short, impersonal, and offers no concrete guidance. We use this interaction throughout this section to illustrate each stage of the pipeline; all numeric values associated with it are illustrative.

\subsection{Problem Definition}

Let $\mathcal{D} = \{(q_i, r_i, m_i, y_i)\}_{i=1}^{n}$
be a dataset of interactions, where $q_i$ is the patient question, $r_i$ is the doctor response, $m_i$ denotes non-textual metadata, and $y_i \in \{0,1\}$ indicates whether the interaction received positive patient feedback. The metadata may include variables such as response time, doctor activity history, and other information available at the time the response is written.

% We address the following question: \textit{Given patient-doctor interactions, what aspects of the doctor response should be minimally changed to maximize the probability of user satisfaction?}

We assume that the doctor response $r_i$ can be characterized by a set of interpretable semantic features $\mathcal{F} = \{f_1, \ldots, f_k\}$
where each feature describes a mutable aspect of communication style, such as empathy or actionability. Each feature $f_j$ takes values from an ordered finite set
$R_j = \{r_{j,1}, \ldots, r_{j,m_j}\}$, for example, from 5-point Likert scale, from "low" to "high". 

Given a question--response pair $(q_i, r_i)$, an \module{Extractor} LM module maps the interaction to an ordinal semantic representation $s_i = E(q_i, r_i; F) \in R_1 \times \cdots \times R_k$. We train a policy model $\pi$ that estimates $\pi(m_i, s_i) = P(y_i = 1 \mid m_i, s_i)$, which is used to select recommendations. For evaluation, to reduce self-confirmation, we separately train auditor models $\alpha_1, \ldots, \alpha_L$ that are not used for recommendation selection.

\subsection{Dataset-Level Feature Discovery via APO}

We instantiate the dataset-level feature discovery framework of \citet{cosma2026automaticpromptoptimizationdatasetlevel} in the telemedicine feedback setting. In this formulation, the prompt does not directly produce a prediction for each input. Instead, it induces a global feature schema $\mathcal{F}_\phi = \{f_1, \ldots, f_k\}$, where $\phi$ denotes the instruction and example context given to the \module{FeatureProposer}. Unlike standard prompt optimization, where the prompt directly improves task predictions, the quality of $\phi$ is determined only indirectly: a good prompt is one that yields features whose extracted values improve downstream feedback prediction while remaining interpretable and actionable.

Given a subset of the training corpus, we optimize a \module{FeatureProposer} LM to generate candidate semantic features of doctor responses. Each feature consists of a name, an ordered value set, and an extraction instruction. We run the \module{FeatureProposer} under several prompt contexts (examples with feedback labels, without labels, and with written patient feedback when available), manually consolidate recurring features across runs, and remove redundant, clinically unsafe or non-actionable ones. The resulting features define the action space for counterfactual recommendations. We show the final feature set in Appendix \ref{sec:appendix-implementation}; this differs from previous work \cite{niculae-etal-2025-dr} in that the selected features are not defined based on organizational or regulatory needs, which can be suboptimal for feedback prediction, but optimized directly for predicting positive feedback.

\subsection{Unsupervised Prompt Refinement}

The initial feature definitions are often too compact to serve directly as reliable extraction prompts, and since we have no gold labels for the semantic features, we cannot run a supervised prompt optimizer over extraction accuracy. Unlike the original formulation of \citet{cosma2026automaticpromptoptimizationdatasetlevel}, where a single \module{Extractor} module handles all features, we instantiate a specialized \module{Extractor} module $E_j$ for each feature $f_j$.

We refine each prompt with a procedure inspired by the grounded instruction proposal stage of MIPRO \cite{opsahl2024optimizing}, omitting its iterative search stage. For each feature, an LM receives the feature definition, its ordinal value set, representative question and response examples, and a task description. It produces a refined extraction prompt that specifies the meaning of each ordinal category, gives decision criteria, and summarizes domain-specific patterns observed in the examples.
We find that refinement improves the reliability of extracted features. Thus, each feature receives its own extractor: $E_j(q_i, r_i; \psi_j) \rightarrow \hat{s}_{i,j}$, where $\psi_j$ is the refined prompt for feature $f_j$. The full semantic representation is obtained by applying all feature-specific extractors: $\hat{s}_i = (E_1(q_i,r_i;\psi_1), \ldots, E_k(q_i,r_i;\psi_k))$.

For the example interaction, the extractors assign, among others, \feature{empathy\_level} = \textit{no\_empathy}, \feature{actionability\_level} = \textit{weakly\_actionable}, and \feature{problems\_addressed} = \textit{partially\_addressed}: the response is a generic referral without context and does not answer whether the patient should be worried.

\subsection{Feedback Estimation}

To train the policy and auditor models, we split the data temporally, holding out the most recent 20\% of interactions as a fixed evaluation set; this choice reflects deployment, where a model trained on past interactions makes recommendations for future ones. The remaining data is divided into training subsets for the policy model and the auditor models, $\mathcal{D} = \mathcal{D}_{\text{policy}} \cup \mathcal{D}_{\text{auditor}} \cup \mathcal{D}_{\text{eval}}$. 
%{\color{orange} Due to data privacy agreements, we cannot disclose the dataset size; however, each partition is sufficient for stable model training.}

For the policy model $\pi$, we used a CatBoost classifier \cite{dorogush2018catboost} trained on question-response metadata and the extracted semantic features. A tree-based model naturally supports mixed categorical, ordinal, and numerical features and provides feature importance estimates that are useful for further analysis.

The auditor models $\alpha_1, \ldots, \alpha_L$ are trained separately from the policy model. They differ from the policy model in training data, feature subsets and inductive bias. In our experiments, we use a \emph{LogisticRegression} (LR) and an \emph{ExplainableBoostingMachines} (EBM) \cite{lou2013accurate} model, and allow the auditors to use surface-level text features (such as LIWC~\cite{duduau2022development}). The purpose of these models is to test whether recommendations chosen by the policy model remain beneficial under alternative satisfaction estimators. Additional details are presented in Appendix \ref{sec:appendix-implementation}.

\subsection{Searching for Counterfactual Recommendations}

For each response, we search for small positive changes to the extracted semantic features: a counterfactual representation $s'_i$ changes one or more ordinal feature values, and features may only move towards values that correspond to a better communication form according to the feature rubric.

For a feature $f_j$, let $\operatorname{rank}_j(s_{i,j})$ denote the ordinal rank of its current value. The cost of changing feature $j$ is the number of ordinal steps between the original and counterfactual value $c_j(s_i, s'_i) = \left| \operatorname{rank}_j(s'_{i,j}) - \operatorname{rank}_j(s_{i,j})\right|$.
The total intervention cost is $C(s_i, s'_i) = \sum_{j=1}^{k} c_j(s_i, s'_i)$. Given a budget $B$, we consider only counterfactuals whose cost is at most $B$: $\mathcal{X}_B(s_i) = \{s'_i : C(s_i, s'_i) \leq B\}$.

Because the semantic features are ordinal and low-dimensional, we enumerate all valid changes within the budget rather than relying on gradient-based or heuristic search \cite{verma2024counterfactual}, and compute the policy model's predicted positive feedback probability $p(s'_i) = \pi(m_i, s'_i)$ for each candidate. We retain only candidates that improve over the original prediction: $\mathcal{I}_B(s_i) = \{s'_i \in \mathcal{X}_B(s_i) : p(s'_i) > p(s_i)\}$. Finally, we select the candidate closest to the ideal point of maximum predicted feedback and zero intervention cost:

\begin{equation}
    s_i^\star = \argmin_{s'_i \in \mathcal{I}_B(s_i)}
    \sqrt{
    (1 - p(s'_i))^2
    +
    \left(
    \frac{C(s_i, s'_i)}{B}
    \right)^2
    }
\end{equation}

In practice, each feature can also receive a modification weight, since some features are easier to edit than others from doctors' perspectives. The output is a set of feature-level recommendations, such as increasing the \feature{actionability} of the response, presented as an interpretable editing target that the doctor may accept, ignore or adapt.

For the example interaction, the policy model assigns the original response a positive feedback probability of $p(s_i) = 0.41$. With a budget $B = 3$, the selected counterfactual raises \feature{empathy\_level} from \textit{no\_empathy} to \textit{moderate\_empathy} (cost 2) and \feature{actionability\_level} from \textit{weakly\_actionable} to \textit{moderately\_actionable} (cost 1), increasing the predicted probability to $p(s_i^\star) = 0.58$. The doctor receives two editing targets: acknowledge the patient's worry, and add at least one concrete next step, for example, what to monitor or when an in-person visit becomes necessary.

\subsection{Auditing Recommendations}
\label{subsec:auditing}

After selecting $s_i^\star$ with the policy model, we evaluate the same counterfactual under each auditor model and take the average change in probability $\Delta(i) = \mathbb{E}_{l<L}[\alpha_\ell(m_i, s_i^\star) - \alpha_\ell(m_i, s_i)]$.

A recommendation is considered robust when it improves the policy prediction and yields non-negative changes across the auditor models. For the running example, the counterfactual selected by the policy model also yields a positive average delta under the auditors, so the recommendation would be considered robust enough to be shown to the doctor. Our use of independent auditor models is relevant for counterfactual recommendations, since interventions selected by one model can exploit that model's particular errors~\citep{kearns2018preventing,kim2019multiaccuracy,hebertjohnson2018multicalibration,upadhyay2021robust,dutta2022robust}.

%% file: sections/4.exps-results.tex
\begin{figure}[hbt!]
    \centering
    \includesvg[width=0.85\linewidth]{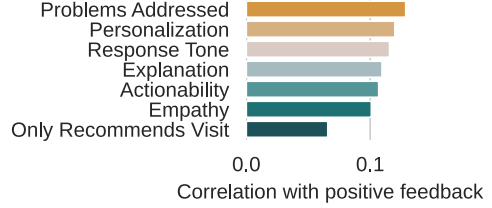}
    \caption{Spearman correlations between extracted features and positive feedback. All features show positive associations with satisfaction.}
    \label{fig:spearman}
\end{figure}
\begin{figure}[hbt!]
    \centering
    \includesvg[width=0.85\linewidth]{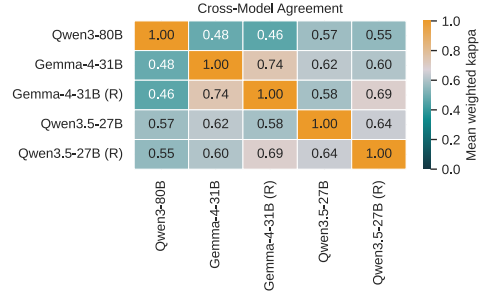}
    \caption{Pairwise agreement between extractor models after majority voting over rollout-level feature values.}
    \label{fig:agreement}
\end{figure}

\paragraph{Evaluating Feature Quality.} In Fig. \ref{fig:spearman} we show Spearman correlations between extracted features and the positive feedback indicator; the extracted features are generally positively associated with positive feedback. Further, in Fig. \ref{fig:agreement} we show agreement among the feature values extracted by the \module{Extractor} LMs. We use \textit{(R)} to denote the refined \module{Extractor} prompt. There is moderate agreement between models; same-family extractor pairs agree more strongly than cross-family pairs, with the \model{Gemma-4-31B} and \model{Gemma-4-31B} \textit{(R)} pair obtaining the highest off-diagonal agreement. Table \ref{tab:feedback_classifier_results} shows the performance of the policy model under various \module{Extractor} models; the discovered semantic features are predictive of patient satisfaction. The \model{Qwen3.5-27B} \textit{(R)} extractor obtains the highest validation ROC-AUC of 0.717, suggesting that the refined semantic annotations yield more effective feature representation in this setting. However, the highest rollout stability score is achieved by the \model{Gemma-4} family. Refining the prompts improves both final downstream performance and rollout stability, at practically negligible cost. We show additional results and implementation details in Appendix \ref{sec:appendix-implementation}.

Consequently, \model{Gemma-4-31B} \textit{(R)} combines near-best downstream performance with the highest rollout stability in Table \ref{tab:feedback_classifier_results}, while remaining highly consistent with its unrefined counterpart. We therefore use \model{Gemma-4-31B} \textit{(R)} as the preferred extractor in subsequent analyses.

\input{tables/feedback_model_clf_metrics}

\begin{figure}[hbt!]
    \centering
    \includesvg[width=0.85\linewidth]{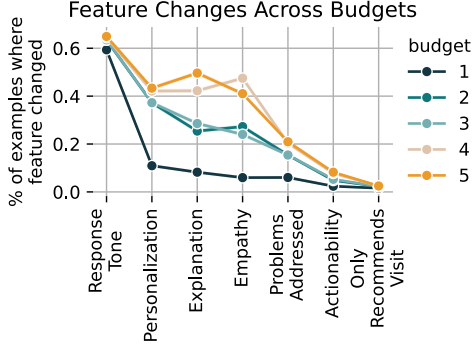}
    \caption{Fraction of recommendations that modify each semantic feature as the edit budget increases.}
    \label{fig:feature-changes}
\end{figure}

\begin{figure}[hbt!]
    \centering
    \includesvg[width=0.85\linewidth]{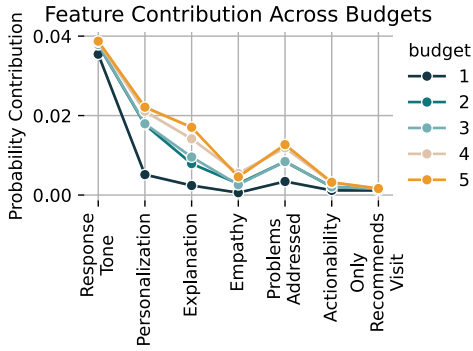}
    \caption{Average probability contribution of each modified feature to the predicted positive-feedback probability across budgets.}
    \label{fig:feature-contribution}
\end{figure}

\paragraph{Evaluating Counterfactual Recommendations.}
In Fig. \ref{fig:feature-changes} we show that increasing the modification budget results in greater diversity among the changeable features. As the budget grows, modifications are spread across a larger set of features, indicating that achieving more substantial interventions requires adjustments to a broader range of attributes. Additionally, in Fig. \ref{fig:feature-contribution} we show that increasing the budget yields only marginal improvements in feature probability contribution. This suggests that relatively minor modifications are already sufficient to achieve most of the performance gains. 

\begin{figure}[hbt!]
    \centering
    \includesvg[width=0.85\linewidth]{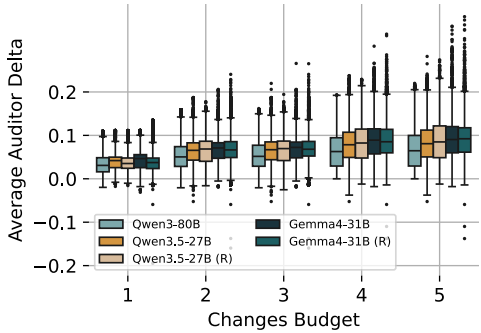}
    \caption{Distribution of average auditor improvement delta across edit budgets and extractor models, after applying the selected counterfactual recommendation. Recommendations selected by the policy model remain beneficial under independent auditor models.}
    \label{fig:audit-delta}
\end{figure}

\input{tables/auditor_results}

In Fig. \ref{fig:audit-delta} we show the distribution of auditor-evaluated deltas across budgets and \module{Extractor} modules. As expected, increasing the change budget results in larger increases in auditor probability estimates, providing further evidence that the proposed method reliably identifies counterfactual recommendations expected to improve user feedback. Additionally, Table \ref{tab:auditor_results}, shows that the recommendation policy obtains strong performance across all auditor configurations: the selected recommendations improved the policy estimate for 95.65\% of responses, with a mean predicted-feedback increase of {+7.35\%}. Under independent auditor models, the same recommendations produced a positive mean delta of {+6.37\%}, and {93.31\%} of deltas were non-negative on average across auditors. We also compare our method with greedy recommendations (see Table \ref{tab:greedy-baseline}, Appendix \ref{sec:appendix-implementation}) and show that our method obtains better improvement rates.

%% file: tables/feedback_model_clf_metrics.tex
\begin{table}[hbt!]
\centering
\resizebox{0.85\linewidth}{!}{
\begin{tabular}{lcc}
\textbf{\module{Extractor}} & \textbf{ROC-AUC} & \textbf{\makecell{Avg. Rollout \\ Stability}} \\
\midrule
\model{Qwen3-80B}     & 0.714 & 0.957 \\
\midrule
\model{Qwen3.5-27B}                     & 0.715 & 0.718 \\
\model{Qwen3.5-27B} \textit{(R)}           & \textbf{0.717} & 0.778 \\
\midrule
\model{Gemma-4-31B}                  & 0.714 & 0.960 \\
\model{Gemma-4-31B} \textit{(R)}        & 0.715 & \textbf{0.974} \\
\end{tabular}
}
\caption{Validation ROC-AUC of the policy model and average rollout stability across the evaluated feature extraction models.}
\label{tab:feedback_classifier_results}
\end{table}

%% file: tables/auditor_results.tex
\begin{table}[hbt!]
\centering
\resizebox{0.85\linewidth}{!}{
\begin{tabular}{llcc}
\textbf{Auditor} & \textbf{\module{Extractor}} & \textbf{\makecell{Improvement \\rate (\%)}} & \textbf{\makecell{Improvement \\delta (\%)}} \\
\midrule
\multirow{5}{*}{EBM} & \model{Qwen3-80B} & 89.58 & +5.2 \\
 & \model{Qwen3.5-27B} & 97.18 & +6.3 \\
 & \model{Qwen3.5-27B} \textit{(R)} & 97.54 & +6.3 \\
 & \model{Gemma-4-31B} & 94.90 & +6.3 \\
 & \model{Gemma-4-31B} \textit{(R)} & \textbf{98.91} & \textbf{+6.6} \\
\midrule
\multirow{5}{*}{LR} & \model{Qwen3-80B} & 89.54 & +5.3 \\
 & \model{Qwen3.5-27B} & 86.25 & +6.4 \\
 & \model{Qwen3.5-27B} \textit{(R)} & 88.68 & +6.7 \\
 & \model{Gemma-4-31B} & 94.71 & \textbf{+7.6} \\
 & \model{Gemma-4-31B} \textit{(R)} & \textbf{95.83} & +7.0 \\
\end{tabular}}
\caption{Counterfactual recommendation performance across auditor models and feature extractors, averaged over edit budgets. Improvement rate is the percentage of responses for which the selected counterfactual yields a non-negative change under the auditor, and improvement delta is the mean change in predicted positive-feedback probability. The best value in each column within an auditor block is shown in \textbf{bold}.}
\label{tab:auditor_results}
\end{table}

%% file: sections/5.conclusions.tex
Our results suggest that \textit{interpretable, low-effort interventions can achieve substantial improvements in predicted feedback, while preserving clinician ownership of medical communication}. Patient feedback was associated with a small number modifiable communication properties, including response tone, personalization and completeness in addressing patient concerns. Counterfactual modifications to these properties remained beneficial when evaluated by independent models that were not involved in selecting them. Most of the predicted improvement was obtained with low editing budgets. This indicates that the practical value of the system lies in identifying the one or two communication deficiencies most worth the clinician's attention. Across responses, the selected recommendations obtained an average improvement of +6.41\% and assigned a non-negative average change to 93.31\% of recommendations. These findings support communication-level counterfactual guidance as a middle ground between passive feedback analytics and automated medical respose generation, as the model identifies actionable communication improvements, while the doctor retains control over the medical reasoning and the final response.

%% file: sections/6.limitations.tex
Although the evaluation is conducted offline, the use of a temporal holdout and independent auditor models provides evidence that the identified recommendations are not artifacts of a single predictive model. However, the proposed approach requires prospective validation in real-world deployment, where doctors apply the recommendations and their impact on patient feedback can be measured.

%% file: sections/7.appendix.tex
\section{Implementation Details}
\label{sec:appendix-implementation}

\paragraph{Models.}
We used \model{Qwen3-80B}~\cite{qwen3technicalreport}, \model{Qwen3.5-27B} \cite{qwen3.5} and \model{Gemma-4-31B}\footnote{\url{hf.co/google/gemma-4-31B-it}, Accessed 16 July, 2026} for semantic feature extraction. These open-weight models provide strong multilingual performance, which is important given that our data is in Romanian, a low-resource language \cite{nigatu2024zeno}. Feature discovery used \model{GPT-5-mini} as the \module{FeatureProposer}, and prompt refinement used \model{GPT-5.4}, both with temperature 1.0 and a 16k output token limit.

\paragraph{Semantic feature extraction.}
Extraction was run offline with vLLM 0.19.1 \cite{kwon2023efficient} on two NVIDIA DGX A100 GPUs, with tensor parallelism across the allocated GPUs, a maximum context length of 16k tokens, and at most 16 concurrent sequences. Each feature is extracted with three rollouts at temperature 1.0, with the final value determined by majority vote over the rollouts.

\input{tables/schema}

\paragraph{Prevalence distribution.}
The prevalence distribution, shown in Table \ref{tab:features} suggests that most responses already exhibit strong communication qualities, including actionability, personalization and adequate coverage of patient concerns. As a result, the recommendation task involves identifying incremental improvements rather than correcting poor responses.

\paragraph{Feedback estimators.}
The policy model is tuned in three stages: SHAP-based \cite{shap} recursive feature selection keeping 10\% of features (at least 32), a 3-fold grid search over tree depth $\{2, 4, 6, 8, 10\}$, learning rate $\{0.01, 0.03, 0.05\}$, and L2 leaf regularization $\{0, 2, 5\}$, and final training for 2500 iterations with balanced class weights. The logistic regression auditor uses imputation, one-hot encoding, and feature scaling. Counterfactual search is run with budgets $B \in \{1, \ldots, 5\}$.

\begin{figure}[hbt!]
    \centering
    \includesvg[width=1.0\linewidth]{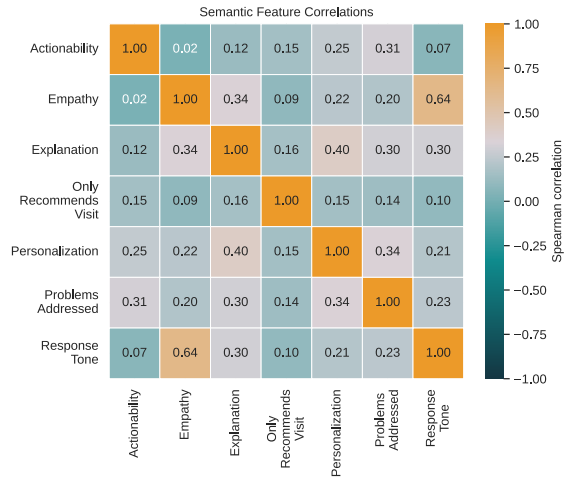}
    \caption{Spearman inter-feature correlations.}
    \label{fig:semantic-features-corr}
\end{figure}

\begin{figure}[hbt!]
    \centering
    \includesvg[width=1.0\linewidth]{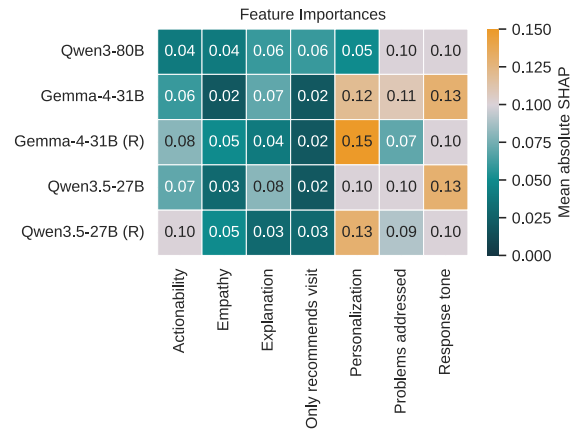}
    \caption{SHAP~\cite{shap} feature importance across \module{Extractor} modules for the policy model in predicting positive feedback.}
    \label{fig:feature-importance}
\end{figure}

\paragraph{Semantic feature correlation.}
Fig. \ref{fig:semantic-features-corr} shows that the extracted features are largely uncorrelated, with the exception of \feature{empathy} and \feature{response\_tone}, which exhibit a moderate positive correlation. This suggests that each feature captures a distinct aspect of response quality and contributes unique information for predicting positive user feedback.

\input{tables/cf_baselines}

\paragraph{Semantic feature importance.}
The feature-importance patterns, depicted in Fig. \ref{fig:feature-importance}, show that across models, the most influential semantic features are \texttt{response\_tone}, \texttt{personalization}, and \texttt{problems\_addressed}. This supports the hypothesis that patient satisfaction is sensitive to the way information is communicated. Notably, these features correspond to aspects that a doctor can revise without delegating medical reasoning to an LM.

\paragraph{Comparison with Greedy Recommendations}
We compare our counterfactual recommendation algorithm with \textit{greedy} recommendations. For a given edit budget, greedy recommendations select the semantic features with the lowest ordinal score and increase its value by the amount specified in the budget. In this case, no policy model is used to guide the selection. In Table \ref{tab:greedy-baseline}, we show the results after applying the auditors; greedy recommendations have lower improvement delta and lower improvement rate overall.

%% file: tables/schema.tex
\begin{table}[hbt!]
    \centering
    \resizebox{\linewidth}{!}{
\begin{tabular}{cl}
\textbf{Feature Name} & \textbf{Values (Prevalence)}\\
\toprule
Actionability & \makecell[l]{unclear\_or\_mixed (0.6\%)\\ not\_actionable (1.7\%)\\ weakly\_actionable (2.7\%)\\ moderately\_actionable (15.9\%)\\ highly\_actionable (79.2\%)} \\
\midrule
Empathy & \makecell[l]{unclear\_or\_mixed (0.4\%)\\ no\_empathy (6.0\%)\\ low\_empathy (58.9\%)\\ moderate\_empathy (30.6\%)\\ high\_empathy (4.1\%)} \\
\midrule
Explanation & \makecell[l]{unclear\_or\_mixed (1.1\%)\\ absent (5.9\%)\\ limited (16.2\%)\\ moderate (26.4\%)\\ clear (50.3\%)} \\
\midrule
Personalization & \makecell[l]{unclear\_or\_mixed (1.7\%)\\ generic (0.3\%)\\ lightly\_personalized (12.3\%)\\ moderately\_personalized (26.3\%)\\ highly\_personalized (59.4\%)} \\
\midrule
Problems Addressed & \makecell[l]{unclear\_or\_mixed (1.1\%)\\ not\_addressed (0.3\%)\\ partially\_addressed (6.0\%)\\ mostly\_addressed (44.6\%)\\ fully\_addressed (47.9\%)} \\
\midrule
Response Tone & \makecell[l]{unclear\_or\_mixed (0.3\%)\\ dismissive\_judgemental (0.4\%)\\ rushed\_curt (0.8\%)\\ neutral\_professional (57.4\%)\\ supportive\_reassuring (41.1\%)}\\
\midrule
Only Recommends Visit & \makecell[l]{yes (2.4\%)\\ no (97.6\%)} \\
\end{tabular}    
    }
    \caption{Extracted features and their corresponding values along with their prevalence.}
    \label{tab:features}
\end{table}

%% file: tables/cf_baselines.tex
\begin{table*}[hbt!]
    \centering
    \resizebox{0.85\linewidth}{!}{
        \begin{tabular}{ll ll | ll}
 &  & \multicolumn{2}{c}{\textbf{Counterfactual Recommendations}} & \multicolumn{2}{c}{\textbf{Greedy Recommendations}} \\
 \textbf{Auditor}  & \textbf{\module{Extractor}} & \textbf{\makecell{Improvement \\rate (\%)}} & \textbf{\makecell{Improvement \\delta (\%)}} & \textbf{\makecell{Improvement \\rate (\%)}} & \textbf{\makecell{Improvement \\delta (\%)}} \\
\midrule
\multirow{5}{*}{EBM} & \model{Qwen3-80B} & 89.58 & +5.3 & 89.49 & +3.4 \\
 & \model{Qwen3.5-27B} & 97.19 & +6.3 & 94.78 & +4.1 \\
 & \model{Qwen3.5-27B} (R) & 97.54 & +6.3 & 98.18 & +4.3 \\
 & \model{Gemma-4-31B} & 94.91 & +6.4 & {99.38} & +3.6 \\
 & \model{Gemma-4-31B} (R) & {98.91} & {+6.6} & 99.23 & {+5.0} \\
\midrule
\multirow{5}{*}{LR} & \model{Qwen3-80B} & 89.55 & +5.3 & {88.29} & +2.2 \\
 & \model{Qwen3.5-27B} & 86.26 & +6.5 & 71.95 & +1.6 \\
 & \model{Qwen3.5-27B} (R) & 88.68 & +6.8 & 74.94 & +2.0 \\
 & \model{Gemma-4-31B} & 94.72 & {+7.6} & 46.98 & +1.0 \\
 & \model{Gemma-4-31B} (R) & {95.84} & +7.0 & 88.23 & {+2.3} \\
\midrule
\textbf{Average} & & \textbf{93.18} & \textbf{+6.41} & 85.14 & +2.95 \\
\end{tabular}
    }
    \caption{Comparison of our policy-guided counterfactual recommendations with a greedy baseline across auditor models and feature extractors, averaged over edit budgets. Improvement rate is the percentage of responses with a non-negative auditor-evaluated change, while improvement delta is the mean change in predicted positive-feedback probability. Our method achieves higher average improvement rate and substantially larger gains than the greedy baseline.}
    \label{tab:greedy-baseline}
\end{table*}